\documentclass[conference]{IEEEtran}
\IEEEoverridecommandlockouts
\usepackage{lineno}
\usepackage{cite}
\usepackage{amsmath,amssymb,amsfonts}
\usepackage{algorithmic}
\usepackage{graphicx}
\usepackage{textcomp}
\usepackage{xcolor}
\usepackage{cleveref}
\usepackage{booktabs}
\usepackage{url}
\usepackage[square,numbers]{natbib}
\usepackage{caption}
\newcommand{\cmark}{\textcolor{green!60!black}{\ding{51}}} 
\newcommand{\xmark}{\textcolor{red}{\ding{55}}} 

\usepackage{graphicx}
\usepackage{epstopdf} 

\usepackage{algorithm}
\usepackage{algorithmic}
\usepackage{amssymb}
\usepackage{pifont}
\usepackage{booktabs}
\usepackage{xcolor}

\usepackage{epstopdf}

\definecolor{darkgreen}{rgb}{0,0.5,0}

\makeatletter
\renewcommand{\paragraph}{%
  \@startsection{paragraph}{4}%
  {\z@}{0.1em}{-1em}%
  {\normalfont\normalsize\bfseries}%
}
\makeatother

\begin{document}

\title{\bfseries \LARGE Universal Neural Architecture Space: Covering ConvNets, Transformers and Everything in Between}

\author{Ondřej Týbl\\
Department of Cybernetics\\
FEE, Czech Technical University\\
{\tt\small tyblondr@cvut.cz}
\and
Lukáš Neumann\\
Department of Cybernetics\\
FEE, Czech Technical University\\
{\tt\small lukas.neumann@cvut.cz}
}

\maketitle

\begin{abstract}
We introduce \textbf{Uni}versal \textbf{N}eural \textbf{A}rchitecture \textbf{S}pace (\textbf{UniNAS}), a generic search space for neural architecture search (NAS) which unifies convolutional networks, transformers, and their hybrid architectures under a single, flexible framework. Our approach enables discovery of novel architectures as well as analyzing existing architectures in a common framework. We also propose a new search algorithm that allows traversing the proposed search space, and demonstrate that the space contains interesting architectures, which, when using identical training setup, outperform state-of-the-art hand-crafted architectures. Finally, a unified toolkit including a standardized training and evaluation protocol is introduced to foster reproducibility and enable fair comparison in NAS research. Overall, this work opens a pathway towards systematically exploring the full spectrum of neural architectures with a unified graph-based NAS perspective.
\end{abstract}
    
\section{Introduction}
\label{sec:intro}

Although neural architecture search (NAS) has achieved undeniable success in identifying optimal hyperparameter configurations for predefined architectures~\cite{nasnet,tan2019rethinking} or in improving inference latency on edge devices~\cite{mnasnet, howard2019searching}, to the best of our knowledge, it has so far failed to produce a novel network architecture that would meaningfully outperform popular yet hand-crafted network architectures based on ResNets~\cite{he2016deep, tan2019efficientnet, tan2021efficientnetv2} or Vision Transformers~\cite{dosovitskiy2020image, dai2021coatnet, tu2022maxvit}. 

As in many areas of computer vision research, NAS advances are primarily driven by available benchmarks, and as such in recent years, the majority of NAS methods~\cite{dudziak2020brp, white2021powerful, ning2022ta, kadlecova2024surprisingly} were focused on tabular NAS benchmarks of the NAS-Bench family~\cite{ying2019bench, dong2020bench}. A tabular benchmark is a fixed dataset of network architectures, where accuracy and other parameters of every network are already pre-computed and therefore the NAS algorithm does not need to train the network in order to obtain its accuracy. While the tabular datasets help to foster NAS research by reducing the required computational costs, by definition they also prevent the NAS methods from discovering new better architectures -- the best architecture of the search space is known already, and one could find it by simply looking up the pre-computed table of accuracies. Moreover, the search space of these tabular datasets is typically relatively small and limited to repeating the same building cell many times, due to the fact that it was necessary to train every network at least once during the benchmark creation.

\begin{figure}[t!]
    \centering
    \includegraphics[width=0.9\linewidth,trim=0 10 0 10, clip]{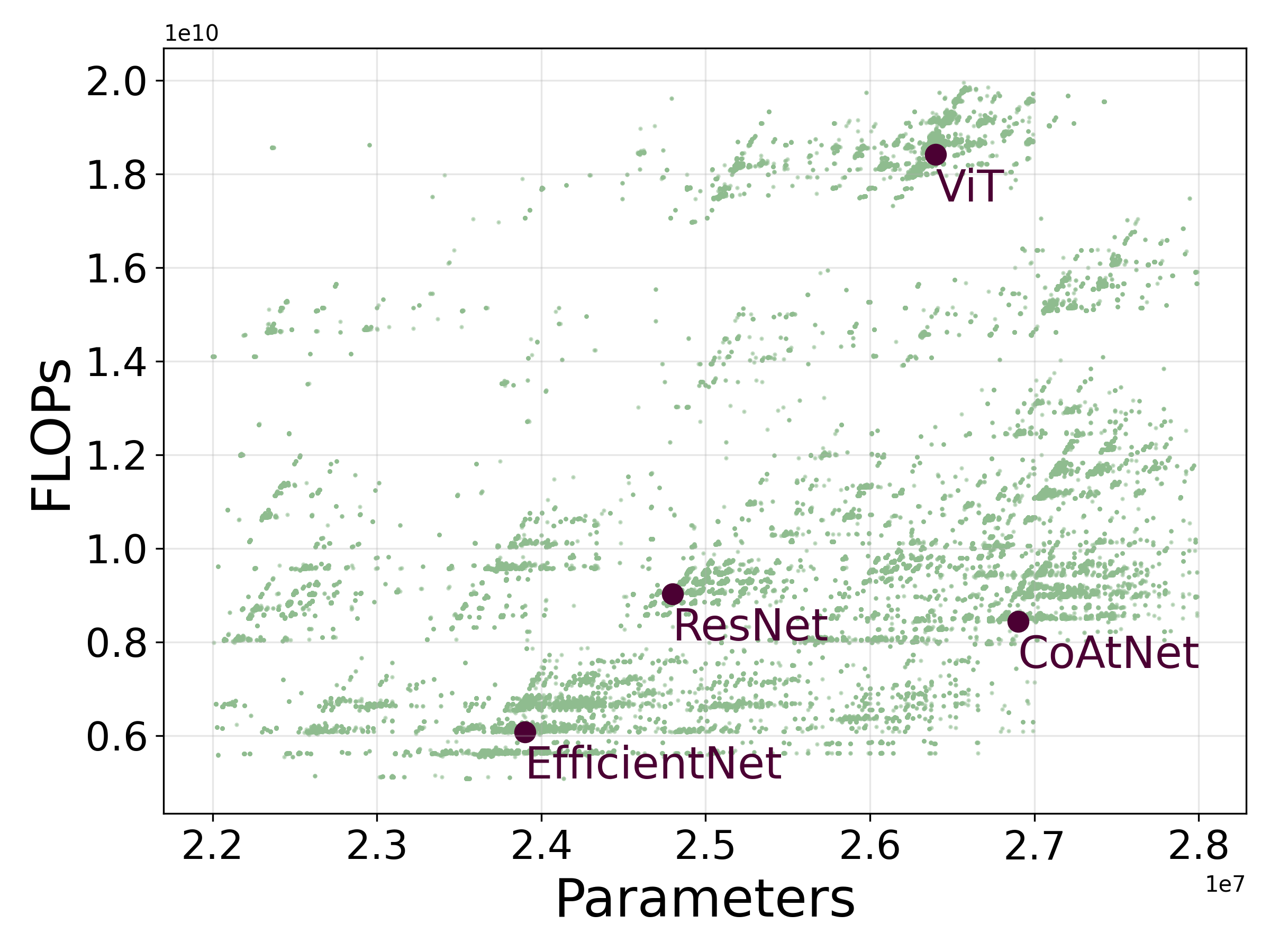}
    \caption{UniNAS search space spans a great variety of different network architectures, including ViT~\cite{dosovitskiy2020image} and EfficientNet~\cite{tan2019efficientnet}, which enables us to search for completely novel architectures. (500,000 networks from a random walk.)}
    \label{fig:figure1}
\end{figure}
\begin{figure*}[t!]
    \centering
    \includegraphics[width=0.8\linewidth, trim=0 14 0 7, clip]{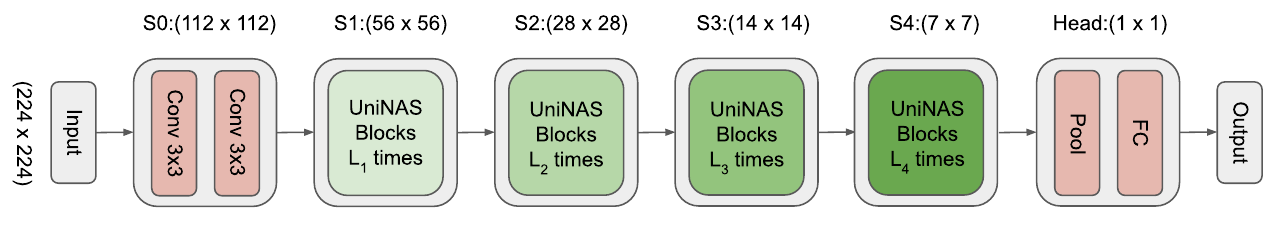}
    \caption{UniNAS architecture. We adopt a standard hierarchical design (e.g., \cite{dai2021coatnet, tu2022maxvit}) with stem, four stages (each with several blocks) and a classification head but introduce a new type of basic building block UniNAS that integrates the universal directed acyclic graph; each of the UniNAS blocks has a \textit{different topology} (see \cref{sec:block}). For clarity, we omit normalization and activation layers.}
    \label{fig:network_structure}
\end{figure*}

Thanks to the emergence of training-free NAS methods~\cite{chen2021neural, zhang2021gradsign, lin2021zen, li2023zico, lee2024az, tybl2025training}, it is possible to estimate the accuracy of a network architecture without training it, thus removing the computational constraint of NAS and enabling exploration of much larger search spaces and novel architectures. The search spaces introduced for this purpose -- namely Zen-NAS~\cite{lin2021zen} and AutoFormer~\cite{chen2021autoformer} -- were created by varying hyperparameters such as depth, width or an expansion ratio of a selected hand-crafted architecture block -- MobileNetV2~\cite{sandler2018mobilenetv2} and Vision Transformer~\cite{dosovitskiy2020image} respectively. Since each search space is created by changing hyperparameters of a given architecture block, it is still impossible to find networks with different topologies or building blocks using these search spaces, and subsequently, hand-crafted architectures such as CoAtNet~\cite{dai2021coatnet}, for instance, still outperform the best known architecture of the AutoFormer search space \cite{casarin2025swag}.

In this paper, we aim to address the above limitations by introducing a novel search space called Universal Neural Architecture Space (\textit{UniNAS}). The search space is designed in a generic manner, without preference towards a particular existing architecture, while at the same time ensuring that state-of-the-art but hand-crafted architectures are all part of the search space. As a result, the search space allows exploration of various combinations of existing architectures, as well as of completely novel architectures, and it allows for systematic analysis of state-of-the-art networks, their topologies and design choices in a unified framework.

Additionally, we propose a new architecture search algorithm to traverse the proposed UniNAS search space. When combined with the state-of-the-art training-free NAS proxy~\cite{tybl2025training}, after only a small number of steps traversing the search space we found a novel network architecture -- UniNAS-A that when using identical training protocol and size constraints outperforms current state-of-the-art architectures, demonstrating that the search space contains interesting architectures and it is worth exploring it further.

Last but not least, we provide a toolkit (as an easily-installable Python package) that allows traversing the proposed search space, creating PyTorch network modules for any point in the search space, and most importantly a consistent and well-defined training protocol and training code to create and evaluate the final model. This is extremely important for reproducibility and for a fair comparison of NAS methods, because while authors would typically report final accuracy on a standard dataset such as ImageNet, they would use different training schedules, different hyperparameter settings and data augmentations, or they would even use a pre-existing larger model trained on significantly more data as the teacher network -- making a fair comparison between proposed architectures virtually impossible.


To summarize, we make the following contributions to the community:
\begin{enumerate}
\item We introduce a novel universal NAS search space called UniNAS. The search space is designed in a generic manner and as such contains many novel network topologies, while at the same time it also contains known hand-crafted architectures, allowing systematic analysis of network topologies in a unified framework.
\item We introduce a new search algorithm to traverse the proposed space and show that it contains interesting novel architectures -- such as UniNAS-A -- which outperforms hand-crafted state-of-the-art architectures on a variety of tasks -- classification, detection and segmentation.
\item We provide a unified toolkit including a standardized training and evaluation protocol to help adoption of the proposed universal search space, to enhance reproducibility of NAS methods and to foster future NAS research.
\end{enumerate}

\section{UniNAS}
\label{sec:uninas}

This section introduces the UniNAS search space, beginning with the UniNAS block as its core component, followed by the overall network architecture, and concluding with a comparison to existing search spaces highlighting key differences and improvements.

\subsection{UniNAS Block}
\label{sec:block}

\begin{table*}[t]
  \centering
  \footnotesize
  \setlength{\tabcolsep}{2pt}
  \begin{tabular}{lccccll}
\toprule
    Node & Input & Output & \texttt{Params} & \texttt{FLOPs} & Description\\
    \midrule
Softmax & $C,H,W$ & $C,H,W$ & $0$ & $C H  (3 W - 1)$ & softmax along the last dim.\\
Dropout & $C,H,W$ & $C,H,W$ & $0$ & $C H W$ & dropout, $p=0.5$\\
MaxPool2d & $C,H,W$ & $C,H,W$ & $0$ & $9 C H  W$ & maxpool, kernel $3$, no stride\\
Mask & $C,H,W$ & $C,H,W$ & $0$ & $C H W$ & keep values up to $5$ pix. from diag.\\
Sigmoid & $C,H,W$ & $C,H,W$ & $0$ & $3 C H  W$ & sigmoid\\
GELU & $C,H,W$ & $C,H,W$ & $0$ & $3 C H  W$ & GELU\\
Conv1 & $C,H,W$ & $C,H,W$ & $C\left(C+1\right)$ & $2 C^2 H W$ & convolution, kernel $1$\\
Conv3 & $C,H,W$ & $C,H,W$ & $C\left(9 C+1\right)$ & $18 C^2 H W$ & convolution, kernel $3$\\
ConvDepth3 & $C,H,W$ & $C,H,W$ & $9 C$ & $18 C H W$ & depthwise convolution, kernel $3$\\
ConvDepth5 & $C,H,W$ & $C,H,W$ & $25 C$ & $50 C H W$ & depthwise convolution, kernel $5$\\
BatchNorm & $C,H,W$ & $C,H,W$ & $2 C$ & $2 C H W$ & batch normalization\\
LayerNorm & $C,H,W$ & $C,H,W$ & $2 C$ & $2 C H W$ & layer normalization\\
RelPosBias & $C,H,W$ & $C,H,W$ & $2(\sqrt{H}-\frac{1}{2})(\sqrt{W}-\frac{1}{2})$ & $C H W$ & add relative position bias\\
Chunk2 & $C,H,W$ & $2\times\left(\frac{C}{2},H,W\right)$ & $0$ & $0$ & chunk into $2$ branches\\
Chunk3 & $C,H,W$ & $3\times\left(\frac{C}{3},H,W\right)$ & $0$ & $0$ & chunk into $3$ branches \\
Copy & $C,H,W$ & $2 \left(C,H,W\right)$ & $0$ & $0$ & copy into $2$ branches \\
Concat2 & $2\times\left(C,H,W\right)$ & $\left(2C,H,W\right)$ & $0$ & $0$ & concat $2$ branches \\
Concat3 & $3\times\left(C,H,W\right)$ & $\left(3C,H,W\right)$ & $0$ & $0$ & concat $3$ branches \\
Add & $2\times\left(C,H,W\right)$ & $\left(C,H,W\right)$ & $0$ & $C H W$ & add $2$ branches \\
ConvChunk3 & $C,H,W$ & $3\times\left( C,H,W\right)$ & $3 C\left(C+1\right)$ & $6 C^2 H W$ & expand channels $\times 3$ and chunk \\
ConvExp4 & $C,H,W$ & $4C,H,W$ & $4 C\left(C+1\right)$ & $8 C^2 H W$ & expand channels $\times 4$\\
ConvRed4 & $C,H,W$ & $\frac{C}{4},H,W$ & $4 C\left(C+1\right)$ & $8 C^2 H W$ & reduce channels $\times 4$ \\
Multiply & $2\times\left(C,H,W\right)$ & $C,H,W$ & $0$ & $4 C H W$ & multiply $2$ branches\\
Matmul1 & $2\times\left(C,H,W\right)$ & $1,H^2,W^2$ & $0$ & $2 C H^2 W^2$ &  see \cref{eq:matmul1}\\
Matmul2 & $\left(C,H^2,W^2\right), \left(C,H,W\right)$ & $C,H,W$ & $0$ & $2 C H^2 W^2$ & see \cref{eq:matmul2}\\
GlobalAvg & $C,H,W$ & $C,1,1$ & $0$ & $C H W$ & global average pooling\\
UpSample & $C,1,1$ & $C,H,W$ & $0$ & $C H W$ & upsample to spatial dimension \\
    \bottomrule
  \end{tabular}
  \caption{UniNAS Block elementary operations. We report number of parameters (\texttt{Params}) and floating point operations (\texttt{FLOPs}) for a single forward pass with input size $C,H,W$. Some nodes change the shape and/or operate in multiple-input or multiple-output regime, $n\times\left(C, H, W\right)$ denotes $n$ tensors of the same shape.}
  \label{tab:nodes}
\end{table*}

\paragraph{Elementary operations.} As our aim is to create a search space that spans a variety of very different architectures, we allow a general setting in the definition of our UniNAS block, enabling maximal flexibility during the architecture search while retaining practical constraints for compatibility with modern hierarchical networks. Specifically, the UniNAS block is defined as any directed acyclic graph (DAG) with one input node and one output node, where each intermediate node represents an elementary operation. The elementary operations include convolution layers (depthwise, pointwise, standard, etc.), pooling, masking, regularization elements, various forms of matrix multiplications and dot products, operations with multiple input or multiple output edges and nonlinearities (see~\cref{tab:nodes}). This allows the block to represent diverse local computation patterns while maintaining consistent interface properties.

While the majority of elementary operations in \cref{tab:nodes} are indeed elementary and therefore do not need further explanation, let us specifically introduce two of the operations which are more complex: for tensors $x, y\in\mathbb{R}^{C\times H\times W}$, we define
\begin{align}
    \texttt{Matmul1}\left(x,y\right)_{h_1,w_1, h_2,w_2}=\frac{1}{\sqrt{C}}\sum_{c=1}^Cx_{c,h_1,w_1}y_{c,h_2,w_2}
    \label{eq:matmul1}
\end{align}
for $h_1,h_2=1,\dots H$ and $w_1,w_2=1,\dots W$ and for $x\in\mathbb{R}^{H^2\times W^2}\simeq\mathbb{R}^{H\times H\times W\times W}, y\in\mathbb{R}^{C\times H\times W}$
\begin{align}
    \texttt{Matmul2}\left(x,y\right)_{c, h,w}=\sum_{\tilde{w}=1}^W\sum_{\tilde{h}=1}^Hx_{h, w, \tilde{h}, \tilde{w}}y_{c, \tilde{h}, \tilde{w}}
    \label{eq:matmul2}
\end{align}
for $h=1,\dots,H$, $w=1,\dots W$ and $c=1,\dots C$. \texttt{Matmul1} and \texttt{Matmul2} represent a multiplicative relationship but unlike the element-wise \texttt{Multiply} it combines all channel information and combination with \texttt{Softmax}, these can be combined to produce the attention mechanism. However, again we note that the variability goes well beyond that as \textit{any} graph is possible and only the dimensions matter (!).

\paragraph{Block computation graph.}
The only constraints we impose on this general formulation are that 1) the input and output dimensions of the block remain identical and that 2) the dimensions between adjacent nodes within the DAG match to ensure correct tensor propagation through the graph. This constraint simplifies integration within larger architectures, where consistent feature map shapes across blocks significantly reduce the complexity of dynamic shape handling during the forward pass and enables stable training across a wide variety of searched blocks. By enforcing consistent dimensions, we avoid the need for automatic insertion of additional projections, which could otherwise interfere with the analysis of the searched topologies.

\begin{table*}[t]
\centering
\small
\setlength{\tabcolsep}{2pt}
\begin{tabular}{lcccccl}
\toprule
 & Block  & Network & ConvNets & Transformers & Novel & Accuracy of the best   \\
 & Topology &  Topology &  &  & architectures & known network (\%)\\
\midrule
DARTS \cite{liu2018darts}  &  \small \color{darkgreen}dynamic &  \small \color{red}repeat same block & \cmark & \xmark & \xmark & $76.1$~\cite{ye2022b} \\
NAS-Bench-101 \cite{ying2019bench}  & \small \color{darkgreen}dynamic  &  \small \color{red}repeat same block &  \cmark & \xmark & \xmark & $75.9$~\cite{liu2020labels} \\
NAS-Bench-201 \cite{dong2020bench}  &  \small\color{darkgreen}dynamic  &  \small \color{red}repeat same block & \cmark & \xmark & \xmark & $76.5$\textsuperscript{\dag} \\
Zen-NAS \cite{lin2021zen}    & \small\color{red} fixed (MBV2~\cite{sandler2018mobilenetv2}) & \small \color{orange} vary block hyper-parameters& \cmark & \xmark & \cmark & 80.1\textsuperscript{\dag} \\
AutoFormer~\cite{chen2021searching} (V2 \cite{chen2021autoformer})   & \small \color{red} fixed (ViT~\cite{dosovitskiy2020image,liu2021swin}) & \small \color{orange} vary block hyper-parameters&  \xmark & \cmark & \cmark & $80.9$\textsuperscript{\dag} \\
\textit{UniNAS (ours)}   & \small \color{darkgreen}dynamic & \small \color{darkgreen} different blocks & \cmark & \cmark & \cmark & \textbf{81.2} \\
\bottomrule
\end{tabular}\vspace{-5pt}
\caption{NAS search spaces comparison. UniNAS spans the largest variety of networks, it does not limit block topology and allows for exploration of novel architectures. We also report the classification accuracy of the best known network of each space on ImageNet-1k -- where we did not find reported results in a comparable training setting, we train the best known architecture of the given search space using the same training protocol as in \cref{tab:hyperparams} (denoted with \textsuperscript{\dag}).
}
\label{tab:spaces}
\end{table*}




We emphasize that many commonly used modules in modern deep learning architectures, such as residual blocks of ResNet~\cite{he2016deep} and its variants, self-attention layers with or without relative position bias as seen in transformer-based models~\cite{dosovitskiy2020image, dai2021coatnet}, squeeze-and-excitation modules~\cite{hu2018squeeze} for adaptive channel recalibration, and inverted mobile bottleneck structures used in EfficientNets~\cite{tan2019efficientnet,tan2021efficientnetv2}, can all be expressed as instances of our UniNAS block. Examples of how such modules can be made into the UniNAS block format, including precise graph structures and node operations, can be seen in \cref{fig:graph_mb}. 

However, we encourage the reader not to stick to these classical examples; one can certainly imagine many more diverse networks. Instead of simply stacking convolutions and nonlinearities in a chain, the search space encompasses tree-like structures, parallel paths with selective attention merging, and hybrid compositions of convolutions and self-attention layers, all encapsulated as a UniNAS block. This demonstrates the expressive capacity of the block design while maintaining a unified and coherent representation throughout the search space.

\subsection{UniNAS Network}
\label{sec:network}

The final network structure sequentially stacks \textit{different UniNAS blocks} into a single chain (see \cref{fig:network_structure}). This follows the general design of the current state-of-the-art networks where the hierarchical backbone is similar to \cite{dai2021coatnet, tu2022maxvit, he2016deep, tan2021efficientnetv2}. The stem stage (S0), where the input is down-sampled using convolution layers, is followed by several stages, each with multiple \textit{different} UniNAS blocks, and finally a classification head consisting of a global average pool and a fully connected layer. As a UniNAS block preserves dimensions, at the beginning of each stage, the spatial dimension is decreased by standard max-pooling with stride two and the number of channels increased by a channel projection\footnote{As suggested by \cite{pennington2018spectrum}, stacking convolutions without nonlinearity in between drastically worsens the network properties and thus we merge channel projection with a first convolution following if there is one. This is in line with classical networks such as \cite{dai2021coatnet}}. As mentioned, the structure (i.e., the graph representation) of the UniNAS blocks varies throughout the network, which enhances the topological variability of the networks. The number of stages, blocks, and the spatial and channel dimensions act as scaling hyperparameters, allowing exploration of different network size modes.

\subsection{Comparison with Existing Search Spaces}
\label{sec:comparison_with_other_spaces}



Existing search spaces suffer from two main limitations: (i) restricted topological variability, and (ii) poor scalability. As a result, the best-performing models discovered within these spaces still underperform compared to architectures obtained with UniNAS, despite the extensive prior work in this area (see \cref{tab:spaces}).

DARTS approaches such as \cite{liu2018darts, zhao2021memory} rely on weight-sharing supernets, which are computationally expensive and yield biased gradient estimates, leading to unreliable architecture rankings. When extended to transformers \cite{chen2021searching}, the cost of self-attention layers forces these search spaces to be heavily constrained—often reducing the search to a binary choice of “use attention or not.” Even with such restrictions DARTS-based networks remain restricted to small setting with ImageNet-1k accuracy below 80\% (\cref{tab:spaces}).

Benchmarks such as NAS-Bench \cite{ying2019bench,dong2020bench} face a fundamental scalability issue: the number of possible networks grows exponentially with the number of operations, making exhaustive training feasible only for toy spaces. Consequently, these benchmarks are limited to simplified convolutional networks evaluated on small datasets like CIFAR~\cite{krizhevsky2009learning} or ImageNet16-120~\cite{to2017adownsampled}. On ImageNet-1k, either no results exist or the reported performance is far below state of the art, with no room for further improvement since the architecture space is finite and already fully explored.

Zen-NAS \cite{lin2021zen} and AutoFormer (V1/V2, also known as S3) \cite{chen2021autoformer,chen2021searching} restrict the search space to networks with repeating MobileNetV2 blocks \cite{sandler2018mobilenetv2} or Vision Transformer blocks \cite{dosovitskiy2020image,liu2021swin}. The resulting architectures differ only by hyper-parameters such as expansion ratios, channel counts, or head numbers. It is however currently impossible to fairly compare top models reported in these spaces, as they are trained through knowledge distillation from a much larger teacher model, using substantially more data~\cite{lee2024az} and it is thus unclear whether reported performance gains are due to the search itself or the distillation procedure. In fact, \cite{faghri2023reinforce} achieved even higher accuracy using a standard EfficientNet-B2 \cite{tan2019efficientnet} under the same FLOPs and training budget, when distillation was applied. Similarly, AutoFormer’s best transformer \cite{chen2021autoformer,casarin2025swag} is outperformed by more than 1\% on ImageNet-1k by a hand-crafted transformer from \cite{xia2023dat++} when trained under the same parameter budget. 

\begin{figure}[t]
    \setlength{\tabcolsep}{2pt}
    \centering
    \begin{tabular}{cc}
         \includegraphics[width=90pt, trim=30 12 38 12, clip]{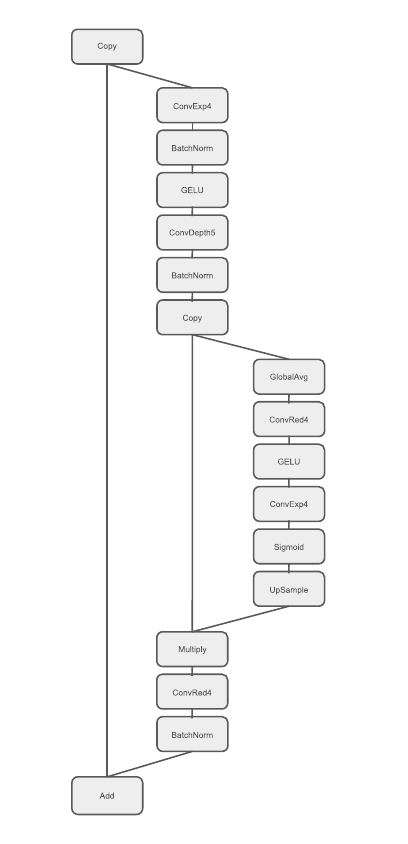} &
         \includegraphics[width=140pt, trim=15 5 15 0, clip]{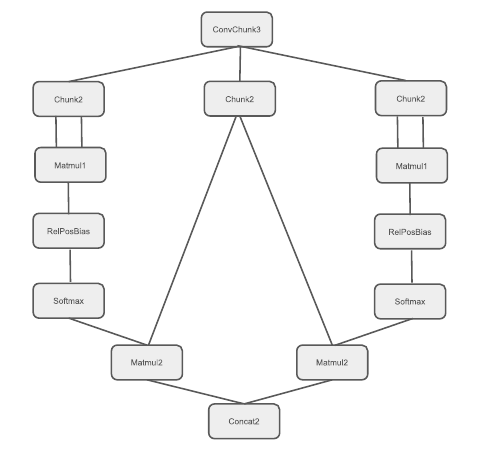}
    \end{tabular}\vspace{-5pt}    
    \caption{UniNAS representation of an EfficientNet (\cite{tan2019rethinking}) block with expansion ratio $4$ (left) and self-attention module (\cite{dosovitskiy2020image}) with $2$ heads (right).}
    \label{fig:graph_mb}
\end{figure}

To summarize, there are two major areas for improvement: network variability and fair reproducible comparison. Both are addressed by UniNAS. 1) flexibility in UniNAS allows for a truly topology-aware architecture search, where attention can be inserted in selected locations, combined with convolutions, or replaced entirely in certain branches, a flexibility that is infeasible in previous frameworks tied to monolithic operation choices; 2) our framework allows for a fair comparison of network topologies as it covers all mentioned spaces and also topologies such as ResNet~\cite{he2016deep}, EfficientNet~\cite{tan2019efficientnet}, ViT~\cite{dosovitskiy2020image}, and CoAtNet~\cite{dai2021coatnet} (see \cref{fig:graph_mb}), so it enables a direct comparison of their properties and accuracy against each other and against novel architectures in one framework.

Unlike previous search spaces, using different blocks at each stage is feasible in our UniNAS (see \cite{lopes2023neural} for a critique on NAS-Bench frameworks stemming from the fact that a simple number of model parameters has the highest prediction strength for estimating final network accuracy, therefore preventing researches from exploring new interesting topologies).

\section{Architecture Search}
\label{sec:architecture_search}

In this Section, we propose an algorithm to find networks in the UniNAS search space based on given criteria. Specifically, we assume that the design choices such as the number of stages, blocks and base channel dimension are known and that the constraints in the form of \texttt{Params} and \texttt{FLOPs} boundaries are given. We then aim to find the network that maximizes certain objective, such as the highest accuracy, given the above constraints. 

\paragraph{Search step.} More formally, any UniNAS network is identified by a sequence of graphs $G_1,\dots,G_{L}$, each graph corresponding to a UniNAS block, where $L$ is the total block count. This allows us to formulate the UniNAS search as a graph-based algorithm involving node additions and eliminations. In order to guarantee a search that efficiently walks through 1) only networks within network size constraints and 2) feasible graphs (in terms of the node dimensionalities and parities), we take the following approach.

1) We associate each node $v\in G_l$, $l=1,\dots,L$ with the number of trainable parameters \texttt{Params} and the approximated number of floating operations \texttt{FLOPs}. These values are easily obtainable as functions of the input tensor shape for each elementary operation node (and can be found in \cref{tab:nodes}), which enables us to efficiently compute the overall network cost after adding/elimination of a given node. 

We note that the \texttt{Params} value for each elementary operation was straightforward to compute. However, estimating \texttt{FLOPs} is non-trivial, and we chose to include all multiplications and additions in the final \texttt{FLOPs} value of each operation. This is so that we give a non-zero cost to all of the operations to prevent an uncontrollable complexity divergence. As a result, the value might be different from values returned by run-time FLOPs estimators like a PyTorch profiler, where, moreover, the value might also differ based on an underlying hardware.

2) We define feasible node addition and elimination operations for our search space. These will guarantee that UniNAS block remains a feasible computation graph after one search step while allowing us to walk freely in the large UniNAS space. Keep in mind that the feasibility of the graph is given by the input/output node shapes in \cref{tab:nodes}.
It is enough to ensure that a) \texttt{RelPosBias} is added only when the input spatial dimension has an integer square root\footnote{Otherwise the operation does not make a mathematical sense; it is motivated by the same module in self-attention (see \cite{dai2021coatnet}).}, b) \texttt{Chunk2, Chunk3} are added only when the channel dimension is divisible by $2$ or $3$ respectively, c) \texttt{ConvRed4} added only when the channel dimension is divisible by $4$, d) nodes changing the dimension and/or having multiple outputs are added together with their \textit{coupled node}. 
When eliminated, the branch between these two is eliminated too.

\begin{algorithm}
\caption{UniNAS Search Step}
\label{alg:search}
\footnotesize
\begin{algorithmic}[1]
\REQUIRE UniNAS network, i.e. a sequence of graphs $G_1,\ldots,G_L$, list of possible node types (cf. \cref{tab:nodes} with node costs \texttt{FLOPs}, \texttt{Params}), search boundaries (\texttt{FLOPs}$_{\max}$, \texttt{FLOPs}$_{\min}$, \texttt{Params}$_{\max}$, \texttt{Params}$_{\min}$), elimination probability $p_{\text{eliminate}} \in [0,1]$, maximum attempts $n_{\text{try}} \in \mathbb{N}, t=0.$
\WHILE{$t < n_{\text{try}}$}
    \STATE Randomly select a graph $G \in \mathcal{G}$ and a node $v \in G$
    \STATE Sample $u \sim \mathcal{U}(0,1)$
    \IF{$u < p_{\text{eliminate}}$} 
        \STATE \COMMENT{Node $v$ elimination}
        \STATE Determine the minimal subgraph $G'$ between $v$ and its coupled node (empty for nodes with one output not changing the dimension)
        \STATE Compute the change in \texttt{FLOPs} and \texttt{Params} after potential elimination of $G'\cup\lbrace v\rbrace$
        \IF{change plausible w.r.t. \texttt{FLOPs}$_{\min}$ and \texttt{Params}$_{\min}$}
            \STATE Eliminate $G'\cup\lbrace v\rbrace$
            \STATE Break
        \ENDIF
    \ELSE
        \STATE \COMMENT{Node addition after $v$}
        \STATE Compute change in \texttt{FLOPs} and \texttt{Params} after potential addition of $v$ and its coupled node(s) $G'$ (empty for nodes with one output not changing the dimension)
        \IF{change plausible w.r.t. \texttt{FLOPs}$_{\max}$ and \texttt{Params}$_{\max}$}
            \STATE Add $G'\cup\lbrace v\rbrace$
            \STATE Break
        \ENDIF
    \ENDIF
    \STATE $t=t+1$
\ENDWHILE
\end{algorithmic}
\end{algorithm}

Finally, we formulate the (single) search step. In each step, we choose a UniNAS block that will be adjusted, decide if we will add or eliminate a node. If we add, we add after a randomly chosen node if it is feasible according to 1) and 2) above. The parameter $p_{\text{eliminate}}$ determining the addition/elimination option is typically set below $0.5$ as we wish to rather add and then eliminate nodes due to the fact that for some types of nodes, their elimination also means elimination of the entire branch between the node and its coupled counterpart, which can result in a large cut in network size. We chose the value $p_{\text{eliminate}}=0.3$ in the experiments as \cref{fig:figure1} suggests that with this choice the random walk avoids degeneracy toward unbounded growth or shrinkage of the network size. For more formal description see \cref{alg:search}.

\begin{figure}[t]
    \centering
    \includegraphics[width=\linewidth, trim=0 325 0 0, clip]{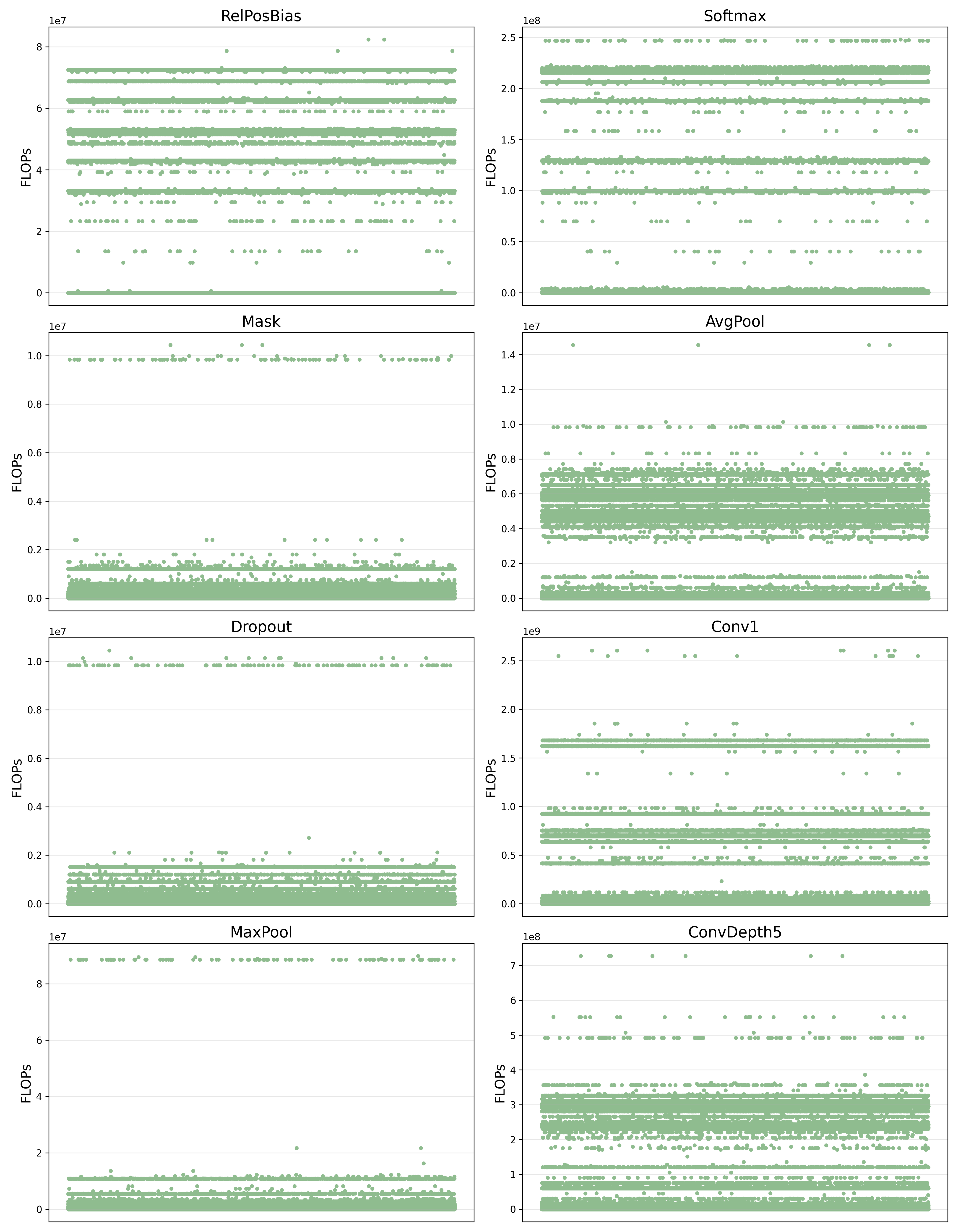}
    \caption{\texttt{FLOPs} per elementary operation for a sample of 500,000 architectures generated via a random walk. A desirable property of the search step in \cref{alg:search} is that it should not collapse into a subspace of degraded networks. Our experiments confirm that the counts for each operation remain both stable and sufficiently variable, indicating that our random walk effectively explores large portions of the UniNAS space.}
    \label{fig:scatters}
\end{figure}

\begin{table}
  \centering
  \footnotesize
  \setlength{\tabcolsep}{1pt}
  \begin{tabular}{lccc}
  \toprule
  & \textbf{Image} & \textbf{Object} & \textbf{Image} \\
  & \textbf{Classification} & \textbf{Detection} & \textbf{Segmentation} \\
    \bottomrule
    training data & ImageNet-1k~\cite{deng2009imagenet} & COCO~\cite{lin2014microsoft} & ADE20K~\cite{zhou2017scene} \\
    network head & FC & Mask R-CNN~\cite{he2017mask} & UperNet~\cite{xiao2018unified} \\
    GPU count & $N$ & $N$ & $N$ \\
    epochs & $150$ & $12$ & $125$ \\
    warmup epochs & $5$ & $5$ & $5$ \\
    batch size  & 48 & 4 & 4\\
    optimizer & AdamW~\cite{loshchilov2017decoupled} & AdamW~\cite{loshchilov2017decoupled} & AdamW~\cite{loshchilov2017decoupled} \\
    weight decay & $0.05$  & $0.05$  & $0.05$  \\
    LR schedule & cosine & multi-step & linear \\
    warmup LR & $N\times 10^{-7}$ & $N\times 10^{-7}$ & $N\times 10^{-7}$ \\
    minimal LR & $N\times 10^{-6}$ & $N\times 2.5\times10^{-6}$ & $0$ \\
    learning rate & $N\times 10^{-4}$ & $N\times 2.5\times 10^{-5}$ & $N\times 1.5\times 10^{-5}$ \\
    data aug. & rand-m15-n2-mstd0.5 & RandFlip0.5 & \begin{tabular}{@{}c@{}}PhotoMetricDist. \\ RandFlip0.5\end{tabular} \\
    gradient clip & $1.0$ & $1.0$ & $1.0$ \\
    drop path & $0.2$ & $0.1$ & $0.3$ \\
    input resolution & $224\times 224$ px & $1280\times 800$ px & $512\times 512$ px \\
    \bottomrule    
  \end{tabular}\vspace{-5pt}
  \caption{UniNAS evaluation protocol. We evaluate NAS methods in the UniNAS space on classification, detection and segmentation tasks by first training image classification on ImageNet-1k and then fine-tuning on COCO and ADE20K datasets with the above hyper-parameters. The batch size is chosen so that it fits into a A100 GPU and is reported \textit{per GPU}, so the learning rate needs to be adjusted accordingly when more GPUs are used.}
  \label{tab:hyperparams}
\end{table}

\paragraph{Training and evaluation protocol.} When a training-free NAS algorithm finishes, the architecture that it had identified as the most promising candidate still needs to be trained to obtain the final (true) accuracy. A precise training recipe has unfortunately been missing in the NAS literature and different authors have been reporting results with different training data, the number of epochs, training batch sizes etc., making direct comparison of different NAS methods impossible. In the UniNAS search space, we therefore also provide a detailed training protocol to train the final network (see \cref{tab:hyperparams}), so that different architectures and NAS methods, including future work, can be compared in a fair and reproducible manner.

\paragraph{UniNAS toolkit.} We provide a pip-installable package, \textit{uninas}, which allows researchers easy access to all components needed to work with the proposed UniNAS space. Specifically, we provide tools for generating any UniNAS network as PyTorch models and an intuitive interface for constructing custom UniNAS networks. In addition, we include a module for the graphical visualization of computation graphs, designed specifically for complex network structures. We also provide an implementation of the search algorithm described in \cref{alg:search}, respecting computation budget constraints, wrapped in a single function call that triggers either a random walk through the UniNAS space or an optimization algorithm with respect to any objective computable from a PyTorch network. Finally, the training protocol described in \cref{tab:hyperparams}, with optional support for distributed training, is included, to promote reproducibility and fair comparison of any future UniNAS networks.
\section{Results}
\label{sec:results}

\paragraph{Random walk in UniNAS.}

We performed a random walk using our search step \cref{alg:search} with \texttt{Params} constrained to $22$–$28$M and \texttt{FLOPs} to $6$–$20$G to evaluate the ability of our architecture search to navigate the gigantic UniNAS space. In \cref{fig:figure1} we can see \texttt{Params} and \texttt{FLOPs} of the 500,000 sampled networks, showing that our search can easily span different network sizes and configurations, also covering very different networks such as transformer- and convolution-based architectures. In \cref{fig:scatters} we provide a further breakdown of UniNAS elementary operations, showing that network sizes, across specific operations, are highly variable, indicating that our exploration effectively covers very diverse configurations.

\paragraph{Architecture search.}

When searching for the best architecture in our UniNAS we use the training-free NAS proxy \textit{VKDNW}~\cite{tybl2025training} to evaluate network performance instead of training it. We define 
\begin{align}
    \text{VKDNW}&= -\sum_{k=1}^9\tilde{\lambda}_k\log\tilde{\lambda}_k, \quad  \tilde{\lambda}_k=\frac{\lambda_k}{\sum_{j=1}^9\lambda_j}
    \label{eq:vkdnw}
\end{align}
where $\lambda_k$ denotes the $k$-th decile of the Empirical Fisher Information Matrix (FIM) eigenvalues as a representation of the FIM spectrum and at initialization under random input batch. Specifically, we computed \cref{eq:vkdnw} for each block $b=1,\dots,L$ separately and then averaged over all of these blocks, resulting in a single scalar value that we maximized for in our search. Our choice of the training-free proxy was motivated by the fact that it is orthogonal to a network size (see Fig. 3 in \cite{tybl2025training}) as our aim is to search for the optimal topology in a specific network size budget.

\begin{figure}
    \centering
    \includegraphics[trim=255 132 260 65, width=0.9\columnwidth, clip]{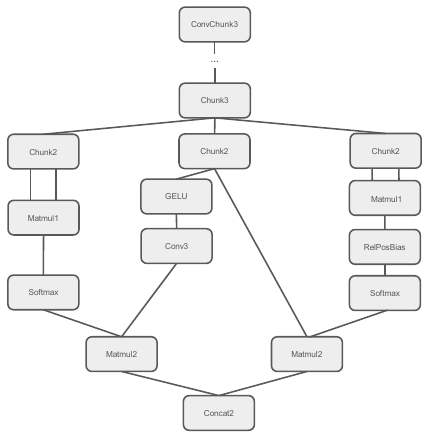}
    \caption{UniNAS-A architecture. The architecture resembles two heads of a self-attention mechanism, but upon closer inspection, certain operations are added or removed.\vspace{-15pt}}
    \label{fig:uninas_architecture}
\end{figure}

We then searched in UniNAS in iterations: we begin with an initial network and perform $1024$ steps from \cref{alg:search}, where only the top $64$ networks (population size) are kept. During the search, the network size was restricted to $27\text{M}$ \texttt{Params} and $20\text{G}$ \texttt{FLOPs} and in \cref{fig:network_structure} we took $4$ stages with stem output size $64$ and stages with $2, 3, 5$ and $2$ UniNAS blocks, respectively, and hidden dimensions $96, 192, 384, 768$ in line with modern architectures\footnote{The batch size for proxy computation \cref{eq:vkdnw} was $64$, the number of steps per iteration, $5$ and $p_{\text{eliminate}}=0.3$ in \cref{alg:search}}. The search took 12 hours on a single A100 GPU and in the end, we chose the best network according to \textit{VKDNW} score -- we denote it as \mbox{\textit{UniNAS-A}} (see \cref{fig:uninas_architecture}).
\begin{table}
\centering
\small
\begin{tabular}{lccc}
\hline
Model & \texttt{Params} & \texttt{FLOPs} & Accuracy (\%) \\
\hline
ResNet \cite{he2016deep} & 24.8M & 9.0G & 75.19 \\
EfficientNet \cite{tan2019efficientnet} & 24.0M & 6.0G & 80.52 \\
ViT \cite{dosovitskiy2020image, dai2021coatnet} & 26.4M & 18.4G & 80.64 \\
CoAtNet \cite{dai2021coatnet} & 26.9M & 8.44G & 80.58 \\
\textit{UniNAS-A (ours)} & 26.8M & 9.0G & \textbf{81.15} \\
\hline
\end{tabular}
\caption{Classification accuracy on ImageNet-1k for models in the UniNAS search space. All models have a similar number of parameters and were trained in an identical protocol, using the training protocol in \cref{tab:hyperparams}. The training took 2 days on 8 A100 GPUs.}
\label{tab:image_classification}
\end{table}

\begin{table*}
\centering
\small
\begin{tabular}{lc|cccc|ccc}
\toprule
& & \multicolumn{4}{c|}{\textbf{Object Detection and Segmentation}} 
& \multicolumn{3}{c}{\textbf{Semantic Segmentation}} \\
\cmidrule(lr){3-6} \cmidrule(lr){7-9}
Model & \texttt{Params} & AP$^b$ & AP$^m$ & FPS (images/s) & FLOPs 
& mIoU & FPS (images/s) & FLOPs \\
\midrule
ResNet \cite{he2016deep}                              & 24.8M & 37.7 & 34.7 & 102.8 & 184G & 39.3 & 481.2 & 47G  \\
EfficientNet \cite{tan2019efficientnet}               & 24.0M & 39.0 & 35.8 & 43.1 & 124G & 37.0 & 152.6 & 32G  \\
CoAtNet \cite{dai2021coatnet}                         & 26.9M & 41.3 & 38.4 & 14.4 & 296G & 42.4 & 61.7 & 51G  \\
\textit{UniNAS-A (ours)}                              & 26.8M & \textbf{42.4} & \textbf{39.0} & 14.2 & 297G & \textbf{45.6} & 88.2 & 51G  \\
\bottomrule
\end{tabular}\vspace{-5pt}
\caption{Accuracy of object detection \& instance segmentation on MS-COCO~\cite{lin2014microsoft} and semantic segmentation on ADE20K~\cite{zhou2017scene} for models in the UniNAS space. $\text{AP}^{\text{b}}$, $\text{AP}^{\text{m}}$ and $\text{mIoU}$ denote box and mask average precision and mean intersection over union, respectively.}
\label{tab:downstream}
\end{table*}


\paragraph{Image Classification.}
First, we compare \textit{UniNAS-A} to other networks in the UniNAS search space: EfficientNet \cite{tan2019efficientnet}, ResNet \cite{he2016deep}, CoAtNet and ViT with relative position bias \cite{dai2021coatnet}. In order to compare the networks in a fair setting, we scaled the networks to the same number of stages and blocks, and by scaling the number of channels, we also adjusted for the same network size. Each network was trained on ImageNet-1k~\cite{deng2009imagenet}, using the same training protocol as in \cref{tab:hyperparams}, the training took 2 days on 8 A100 GPUs. As shown in \cref{tab:image_classification}, \textit{UniNAS-A} outperforms standard hand-crafted networks by a great margin.  

\paragraph{Downstream Tasks.}
We then fine-tune each network from the previous step on two downstream tasks -- object detection on MS-COCO~\cite{lin2014microsoft} and semantic segmentation on ADE20K~\cite{zhou2017scene}, using identical settings from \cref{tab:hyperparams}. Training required approximately 10 hours for segmentation and 21 hours for detection\footnote{To adapt the model to different input sizes, we interpolated the relative position bias module without any other architectural modifications. ViT was excluded from downstream evaluation, as it did not fit to GPU memory with the larger image resolution required by the downstream tasks.} on 4 A100 GPUs. Once again, \textit{UniNAS-A} significantly outperforms existing networks (see~\cref{tab:downstream}).

\paragraph{Proxy Ablation.} Finally, we ablate the choice of VKDNW~\cite{tybl2025training} as our proxy, by running the search \cref{alg:search} with different proxies, and then evaluating the best architecture found by each proxy.  In~\cref{tab:proxy_ablation} we show that the search through the UniNAS space remains mostly robust to the choice of a specific proxy, but VKDNW still being able to find the best network architecture under identical search budget (12 hours).

\begin{table}
\centering
\small
\begin{tabular}{lcccc}
\hline
NAS method & \texttt{Params} & \texttt{FLOPs} & Accuracy (\%) \\
\hline
ZenNAS \cite{lin2021zen} & 26.3M & 18.6G & 80.48 \\
ZiCO \cite{li2023zico} & 27.0M & 9.7G & 80.85 \\
AZ-NAS \cite{lee2024az} & 26.7M & 19.0G & 80.39 \\
VKDNW \cite{tybl2025training} & 26.8M & 9.0G & \textbf{81.15} \\
\hline
\end{tabular}
\caption{Ablation of image classification accuracy on ImageNet-1k in the UniNAS search space, depending on training-free NAS method used. All methods had identical search budget of 12 hours.}
\label{tab:proxy_ablation}
\end{table}

\section{Related Work}
\label{sec:related}
\vspace{-5pt}
As the discussion of NAS search spaces was presented in \cref{sec:comparison_with_other_spaces}, here we focus on NAS search methods.

\paragraph{One-shot NAS.} These methods are based on relaxation of the discrete architecture space, typically using a supernet which contains all possible nodes and edges of the given search space. This approach was first introduced in DARTS~\cite{liu2018darts}, where the supernet consists of all possible operations, each with an assigned weight which is adjusted during supernet training via gradient descent. Once the supernet training is complete, the operation with the highest weight is kept, creating the final network architecture. Robust-DARTS~\cite{zela2019understanding} improves test-time generalization by incorporating data augmentations in supernet training, SGAS~\cite{li2020sgas} on the other hand aims to improve supernet training stability by explicitly selecting allowed operations in each stage. The main challenges of one-shot NAS are memory consumption, because the supernet has to contain all possible operations of the whole search space, and ranking disorder, where the performance of an architecture evaluated in the supernet might be different from the performance of a standalone network. 

\paragraph{Zero-shot NAS.} The methods in the first category aim to create a model that predicts the final accuracy of the network, typically by analyzing the graph structure of the network computation. BRP-NAS~\cite{dudziak2020brp} trains a graph convolutional network to predict network accuracy and latency, while NASLib~\cite{white2021powerful} presents a library of 33 predictors for network accuracy. TA-GATES~\cite{ning2022ta} introduced a novel architecture encoding scheme and GRAF~\cite{kadlecova2024surprisingly} propose new neural graph features. However, all methods in this category require a tabular NAS dataset of architectures and their accuracies to train the model, which makes their application impractical for large or even infinite search spaces.

Methods in the second category try to estimate their accuracy by executing a single forward and backward pass of the network and then by measuring certain properties of the network and/or of the optimization process. Initially, the methods~\cite{tanaka2020pruning, wang2020picking, lee2018snip, abdelfattah2021zero} looked at weight gradients after the first optimization step. This is further improved in Jacov \cite{abdelfattah2021zero} where Jacobian matrices among inputs are compared and in NASWOT~\cite{mellor2021neural} where linear maps between data points are exploited. Zen-NAS~\cite{lin2021zen} then uses the gradient with respect to feature maps and GradSign~\cite{zhang2021gradsign} looks at the optimization landscape of individual training samples. ZiCo~\cite{li2023zico} aggregates gradients from multiple forward and backward passes, TE-NAS~\cite{chen2021neural} combines the number of linear regions with the condition number of Neural Tangent Kernels, AZ-NAS~\cite{lee2024az} then introduces expressivity, trainability and progressivity measures. Most recently, VKDNW~\cite{tybl2025training} uses Fisher Information to characterize the difficulty in estimating the parameters. In these methods, the aim is to rank the networks to obtain the most promising candidates in NAS for the next step of the search process or for the final training.

\vspace{-7pt}
\section{Conclusion}
\label{sec:conclusion}
\vspace{-3pt}
We present UniNAS, a universal neural architecture search space designed to systematically explore, analyze, and compare network topologies within a unified framework. In contrast to previous work, we further decomposed computational modules into elementary operations, enabling the expression and extension of both handcrafted and NAS-discovered architectures while supporting systematic studies of topological variability in network design.

We proposed an efficient architecture search algorithm that operates directly within UniNAS, providing precise control over FLOPs and parameter budgets while allowing fine-grained modifications to traverse diverse architecture families, including convolutional, transformer-based, and hybrid models, within a unified and fair comparison protocol.

We demonstrated that UniNAS contains novel architectures that outperform state-of-the-art hand-crafted networks, underscoring the need for systematic topology search. We provide a unified framework for the search and training of final architectures to facilitate fair future comparisons, made available through a publicly released toolkit.

%

{\small
    \bibliographystyle{ieeenat_fullname}
    \bibliography{main}

\begin{thebibliography}{51}
\providecommand{\natexlab}[1]{#1}
\providecommand{\url}[1]{\texttt{#1}}
\expandafter\ifx\csname urlstyle\endcsname\relax
  \providecommand{\doi}[1]{doi: #1}\else
  \providecommand{\doi}{doi: \begingroup \urlstyle{rm}\Url}\fi

\bibitem[Abdelfattah et~al.(2021)Abdelfattah, Mehrotra, Dudziak, and Lane]{abdelfattah2021zero}
Mohamed~S Abdelfattah, Abhinav Mehrotra, {\L}ukasz Dudziak, and Nicholas~D Lane.
\newblock Zero-cost proxies for lightweight {NAS}.
\newblock \emph{arXiv preprint arXiv:2101.08134}, 2021.

\bibitem[Casarin et~al.(2025)Casarin, Escalera, and Lanz]{casarin2025swag}
Sofia Casarin, Sergio Escalera, and Oswald Lanz.
\newblock L-swag: Layer-sample wise activation with gradients information for zero-shot nas on vision transformers.
\newblock In \emph{Proceedings of the Computer Vision and Pattern Recognition Conference}, pages 4441--4451, 2025.

\bibitem[Chen et~al.(2021{\natexlab{a}})Chen, Peng, Fu, and Ling]{chen2021autoformer}
Minghao Chen, Houwen Peng, Jianlong Fu, and Haibin Ling.
\newblock Autoformer: Searching transformers for visual recognition.
\newblock In \emph{Proceedings of the IEEE/CVF international conference on computer vision}, pages 12270--12280, 2021{\natexlab{a}}.

\bibitem[Chen et~al.(2021{\natexlab{b}})Chen, Wu, Ni, Peng, Liu, Fu, Chao, and Ling]{chen2021searching}
Minghao Chen, Kan Wu, Bolin Ni, Houwen Peng, Bei Liu, Jianlong Fu, Hongyang Chao, and Haibin Ling.
\newblock Searching the search space of vision transformer.
\newblock \emph{Advances in Neural Information Processing Systems}, 34:\penalty0 8714--8726, 2021{\natexlab{b}}.

\bibitem[Chen et~al.(2021{\natexlab{c}})Chen, Gong, and Wang]{chen2021neural}
Wuyang Chen, Xinyu Gong, and Zhangyang Wang.
\newblock Neural architecture search on imagenet in four gpu hours: A theoretically inspired perspective.
\newblock \emph{arXiv preprint arXiv:2102.11535}, 2021{\natexlab{c}}.

\bibitem[Chrabaszcz et~al.(2017)Chrabaszcz, Loshchilov, and Hutter]{to2017adownsampled}
Patryk Chrabaszcz, Ilya Loshchilov, and Frank Hutter.
\newblock A downsampled variant of imagenet as an alternative to the cifar datasets.
\newblock \emph{arXiv preprint arXiv:1707.08819}, 2017.

\bibitem[Dai et~al.(2021)Dai, Liu, Le, and Tan]{dai2021coatnet}
Zihang Dai, Hanxiao Liu, Quoc~V Le, and Mingxing Tan.
\newblock Coatnet: Marrying convolution and attention for all data sizes.
\newblock \emph{Advances in neural information processing systems}, 34:\penalty0 3965--3977, 2021.

\bibitem[Deng et~al.(2009)Deng, Dong, Socher, Li, Li, and Fei-Fei]{deng2009imagenet}
Jia Deng, Wei Dong, Richard Socher, Li-Jia Li, Kai Li, and Li Fei-Fei.
\newblock Imagenet: A large-scale hierarchical image database.
\newblock In \emph{2009 IEEE conference on computer vision and pattern recognition}, pages 248--255. Ieee, 2009.

\bibitem[Dong and Yang(2020)]{dong2020bench}
Xuanyi Dong and Yi Yang.
\newblock Nas-bench-201: Extending the scope of reproducible neural architecture search.
\newblock \emph{arXiv preprint arXiv:2001.00326}, 2020.

\bibitem[Dosovitskiy et~al.(2020)Dosovitskiy, Beyer, Kolesnikov, Weissenborn, Zhai, Unterthiner, Dehghani, Minderer, Heigold, Gelly, et~al.]{dosovitskiy2020image}
Alexey Dosovitskiy, Lucas Beyer, Alexander Kolesnikov, Dirk Weissenborn, Xiaohua Zhai, Thomas Unterthiner, Mostafa Dehghani, Matthias Minderer, Georg Heigold, Sylvain Gelly, et~al.
\newblock An image is worth 16x16 words: Transformers for image recognition at scale.
\newblock \emph{arXiv preprint arXiv:2010.11929}, 2020.

\bibitem[Dudziak et~al.(2020)Dudziak, Chau, Abdelfattah, Lee, Kim, and Lane]{dudziak2020brp}
Lukasz Dudziak, Thomas Chau, Mohamed Abdelfattah, Royson Lee, Hyeji Kim, and Nicholas Lane.
\newblock Brp-nas: Prediction-based nas using gcns.
\newblock \emph{Advances in neural information processing systems}, 33:\penalty0 10480--10490, 2020.

\bibitem[Faghri et~al.(2023)Faghri, Pouransari, Mehta, Farajtabar, Farhadi, Rastegari, and Tuzel]{faghri2023reinforce}
Fartash Faghri, Hadi Pouransari, Sachin Mehta, Mehrdad Farajtabar, Ali Farhadi, Mohammad Rastegari, and Oncel Tuzel.
\newblock Reinforce data, multiply impact: Improved model accuracy and robustness with dataset reinforcement.
\newblock In \emph{Proceedings of the IEEE/CVF International Conference on Computer Vision}, pages 17032--17043, 2023.

\bibitem[He et~al.(2016)He, Zhang, Ren, and Sun]{he2016deep}
Kaiming He, Xiangyu Zhang, Shaoqing Ren, and Jian Sun.
\newblock Deep residual learning for image recognition.
\newblock In \emph{Proceedings of the IEEE conference on computer vision and pattern recognition}, pages 770--778, 2016.

\bibitem[He et~al.(2017)He, Gkioxari, Doll{\'a}r, and Girshick]{he2017mask}
Kaiming He, Georgia Gkioxari, Piotr Doll{\'a}r, and Ross Girshick.
\newblock Mask r-cnn.
\newblock In \emph{Proceedings of the IEEE international conference on computer vision}, pages 2961--2969, 2017.

\bibitem[Howard et~al.(2019)Howard, Sandler, Chu, Chen, Chen, Tan, Wang, Zhu, Pang, Vasudevan, et~al.]{howard2019searching}
Andrew Howard, Mark Sandler, Grace Chu, Liang-Chieh Chen, Bo Chen, Mingxing Tan, Weijun Wang, Yukun Zhu, Ruoming Pang, Vijay Vasudevan, et~al.
\newblock Searching for mobilenetv3.
\newblock In \emph{Proceedings of the IEEE/CVF international conference on computer vision}, pages 1314--1324, 2019.

\bibitem[Hu et~al.(2018)Hu, Shen, and Sun]{hu2018squeeze}
Jie Hu, Li Shen, and Gang Sun.
\newblock Squeeze-and-excitation networks.
\newblock In \emph{Proceedings of the IEEE conference on computer vision and pattern recognition}, pages 7132--7141, 2018.

\bibitem[Kadlecov{\'a} et~al.(2024)Kadlecov{\'a}, Lukasik, Pil{\'a}t, Vidnerov{\'a}, Safari, Neruda, and Hutter]{kadlecova2024surprisingly}
Gabriela Kadlecov{\'a}, Jovita Lukasik, Martin Pil{\'a}t, Petra Vidnerov{\'a}, Mahmoud Safari, Roman Neruda, and Frank Hutter.
\newblock Surprisingly strong performance prediction with neural graph features.
\newblock \emph{arXiv preprint arXiv:2404.16551}, 2024.

\bibitem[Krizhevsky et~al.(2009)Krizhevsky, Hinton, et~al.]{krizhevsky2009learning}
Alex Krizhevsky, Geoffrey Hinton, et~al.
\newblock Learning multiple layers of features from tiny images.
\newblock 2009.

\bibitem[Lee and Ham(2024)]{lee2024az}
Junghyup Lee and Bumsub Ham.
\newblock Az-nas: Assembling zero-cost proxies for network architecture search.
\newblock In \emph{Proceedings of the IEEE/CVF Conference on Computer Vision and Pattern Recognition}, pages 5893--5903, 2024.

\bibitem[Lee et~al.(2018)Lee, Ajanthan, and Torr]{lee2018snip}
Namhoon Lee, Thalaiyasingam Ajanthan, and Philip~HS Torr.
\newblock Snip: Single-shot network pruning based on connection sensitivity.
\newblock \emph{arXiv preprint arXiv:1810.02340}, 2018.

\bibitem[Li et~al.(2020)Li, Qian, Delgadillo, Muller, Thabet, and Ghanem]{li2020sgas}
Guohao Li, Guocheng Qian, Itzel~C Delgadillo, Matthias Muller, Ali Thabet, and Bernard Ghanem.
\newblock {SGAS}: Sequential greedy architecture search.
\newblock In \emph{Proceedings of the IEEE/CVF conference on computer vision and pattern recognition}, pages 1620--1630, 2020.

\bibitem[Li et~al.(2023)Li, Yang, Bhardwaj, and Marculescu]{li2023zico}
Guihong Li, Yuedong Yang, Kartikeya Bhardwaj, and Radu Marculescu.
\newblock {ZiCo}: Zero-shot {NAS} via inverse coefficient of variation on gradients.
\newblock In \emph{The Eleventh International Conference on Learning Representations}, 2023.

\bibitem[Lin et~al.(2021)Lin, Wang, Sun, Chen, Sun, Qian, Li, and Jin]{lin2021zen}
Ming Lin, Pichao Wang, Zhenhong Sun, Hesen Chen, Xiuyu Sun, Qi Qian, Hao Li, and Rong Jin.
\newblock Zen-nas: A zero-shot nas for high-performance image recognition.
\newblock In \emph{Proceedings of the IEEE/CVF international conference on computer vision}, pages 347--356, 2021.

\bibitem[Lin et~al.(2014)Lin, Maire, Belongie, Hays, Perona, Ramanan, Doll{\'a}r, and Zitnick]{lin2014microsoft}
Tsung-Yi Lin, Michael Maire, Serge Belongie, James Hays, Pietro Perona, Deva Ramanan, Piotr Doll{\'a}r, and C~Lawrence Zitnick.
\newblock Microsoft coco: Common objects in context.
\newblock In \emph{European conference on computer vision}, pages 740--755. Springer, 2014.

\bibitem[Liu et~al.(2020)Liu, Doll{\'a}r, He, Girshick, Yuille, and Xie]{liu2020labels}
Chenxi Liu, Piotr Doll{\'a}r, Kaiming He, Ross Girshick, Alan Yuille, and Saining Xie.
\newblock Are labels necessary for neural architecture search?
\newblock In \emph{European Conference on Computer Vision}, pages 798--813. Springer, 2020.

\bibitem[Liu et~al.(2018)Liu, Simonyan, and Yang]{liu2018darts}
Hanxiao Liu, Karen Simonyan, and Yiming Yang.
\newblock {DARTS}: Differentiable architecture search.
\newblock \emph{arXiv preprint arXiv:1806.09055}, 2018.

\bibitem[Liu et~al.(2021)Liu, Lin, Cao, Hu, Wei, Zhang, Lin, and Guo]{liu2021swin}
Ze Liu, Yutong Lin, Yue Cao, Han Hu, Yixuan Wei, Zheng Zhang, Stephen Lin, and Baining Guo.
\newblock Swin transformer: Hierarchical vision transformer using shifted windows.
\newblock In \emph{Proceedings of the IEEE/CVF international conference on computer vision}, pages 10012--10022, 2021.

\bibitem[Lopes et~al.(2023)Lopes, Degardin, and Alexandre]{lopes2023neural}
Vasco Lopes, Bruno Degardin, and Luis~A Alexandre.
\newblock Are neural architecture search benchmarks well designed? a deeper look into operation importance.
\newblock \emph{Information Sciences}, 650:\penalty0 119695, 2023.

\bibitem[Loshchilov and Hutter(2017)]{loshchilov2017decoupled}
Ilya Loshchilov and Frank Hutter.
\newblock Decoupled weight decay regularization.
\newblock \emph{arXiv preprint arXiv:1711.05101}, 2017.

\bibitem[Mellor et~al.(2021)Mellor, Turner, Storkey, and Crowley]{mellor2021neural}
Joe Mellor, Jack Turner, Amos Storkey, and Elliot~J Crowley.
\newblock Neural architecture search without training.
\newblock In \emph{International conference on machine learning}, pages 7588--7598. PMLR, 2021.

\bibitem[Ning et~al.(2022)Ning, Zhou, Zhao, Zhao, Deng, Tang, Liang, Yang, and Wang]{ning2022ta}
Xuefei Ning, Zixuan Zhou, Junbo Zhao, Tianchen Zhao, Yiping Deng, Changcheng Tang, Shuang Liang, Huazhong Yang, and Yu Wang.
\newblock {TA-GATES}: An encoding scheme for neural network architectures.
\newblock \emph{Advances in Neural Information Processing Systems}, 35:\penalty0 32325--32339, 2022.

\bibitem[Pennington and Worah(2018)]{pennington2018spectrum}
Jeffrey Pennington and Pratik Worah.
\newblock The spectrum of the fisher information matrix of a single-hidden-layer neural network.
\newblock \emph{Advances in neural information processing systems}, 31, 2018.

\bibitem[Sandler et~al.(2018)Sandler, Howard, Zhu, Zhmoginov, and Chen]{sandler2018mobilenetv2}
Mark Sandler, Andrew Howard, Menglong Zhu, Andrey Zhmoginov, and Liang-Chieh Chen.
\newblock Mobilenetv2: Inverted residuals and linear bottlenecks.
\newblock In \emph{Proceedings of the IEEE conference on computer vision and pattern recognition}, pages 4510--4520, 2018.

\bibitem[Tan and Le(2019)]{tan2019efficientnet}
Mingxing Tan and Quoc Le.
\newblock Efficientnet: Rethinking model scaling for convolutional neural networks.
\newblock In \emph{International conference on machine learning}, pages 6105--6114. PMLR, 2019.

\bibitem[Tan and Le(2021)]{tan2021efficientnetv2}
Mingxing Tan and Quoc Le.
\newblock Efficientnetv2: Smaller models and faster training.
\newblock In \emph{International conference on machine learning}, pages 10096--10106. PMLR, 2021.

\bibitem[Tan et~al.(2019{\natexlab{a}})Tan, Chen, Pang, Vasudevan, Sandler, Howard, and Le]{mnasnet}
Mingxing Tan, Bo Chen, Ruoming Pang, Vijay Vasudevan, Mark Sandler, Andrew Howard, and Quoc~V Le.
\newblock Mnasnet: Platform-aware neural architecture search for mobile.
\newblock In \emph{Proceedings of the IEEE/CVF conference on computer vision and pattern recognition}, pages 2820--2828, 2019{\natexlab{a}}.

\bibitem[Tan et~al.(2019{\natexlab{b}})Tan, Le, et~al.]{tan2019rethinking}
Mingxing Tan, Q~Efficientnet Le, et~al.
\newblock Rethinking model scaling for convolutional neural networks.
\newblock In \emph{Proceedings of the International conference on machine learning, Long Beach, CA, USA}, 2019{\natexlab{b}}.

\bibitem[Tanaka et~al.(2020)Tanaka, Kunin, Yamins, and Ganguli]{tanaka2020pruning}
Hidenori Tanaka, Daniel Kunin, Daniel~L Yamins, and Surya Ganguli.
\newblock Pruning neural networks without any data by iteratively conserving synaptic flow.
\newblock \emph{Advances in neural information processing systems}, 33:\penalty0 6377--6389, 2020.

\bibitem[Tu et~al.(2022)Tu, Talebi, Zhang, Yang, Milanfar, Bovik, and Li]{tu2022maxvit}
Zhengzhong Tu, Hossein Talebi, Han Zhang, Feng Yang, Peyman Milanfar, Alan Bovik, and Yinxiao Li.
\newblock Maxvit: Multi-axis vision transformer.
\newblock In \emph{European conference on computer vision}, pages 459--479. Springer, 2022.

\bibitem[Tybl and Neumann(2025)]{tybl2025training}
Ondrej Tybl and Lukas Neumann.
\newblock Training-free neural architecture search through variance of knowledge of deep network weights.
\newblock In \emph{Proceedings of the Computer Vision and Pattern Recognition Conference}, pages 14881--14890, 2025.

\bibitem[Wang et~al.(2020)Wang, Zhang, and Grosse]{wang2020picking}
Chaoqi Wang, Guodong Zhang, and Roger Grosse.
\newblock Picking winning tickets before training by preserving gradient flow.
\newblock \emph{arXiv preprint arXiv:2002.07376}, 2020.

\bibitem[White et~al.(2021)White, Zela, Ru, Liu, and Hutter]{white2021powerful}
Colin White, Arber Zela, Robin Ru, Yang Liu, and Frank Hutter.
\newblock How powerful are performance predictors in neural architecture search?
\newblock \emph{Advances in neural information processing systems}, 34:\penalty0 28454--28469, 2021.

\bibitem[Xia et~al.(2023)Xia, Pan, Song, Li, and Huang]{xia2023dat++}
Zhuofan Xia, Xuran Pan, Shiji Song, Li~Erran Li, and Gao Huang.
\newblock Dat++: Spatially dynamic vision transformer with deformable attention.
\newblock \emph{arXiv preprint arXiv:2309.01430}, 2023.

\bibitem[Xiao et~al.(2018)Xiao, Liu, Zhou, Jiang, and Sun]{xiao2018unified}
Tete Xiao, Yingcheng Liu, Bolei Zhou, Yuning Jiang, and Jian Sun.
\newblock Unified perceptual parsing for scene understanding.
\newblock In \emph{Proceedings of the European conference on computer vision (ECCV)}, pages 418--434, 2018.

\bibitem[Ye et~al.(2022)Ye, Li, Li, Chen, Fan, and Ouyang]{ye2022b}
Peng Ye, Baopu Li, Yikang Li, Tao Chen, Jiayuan Fan, and Wanli Ouyang.
\newblock b-darts: Beta-decay regularization for differentiable architecture search.
\newblock In \emph{proceedings of the IEEE/CVF conference on computer vision and pattern recognition}, pages 10874--10883, 2022.

\bibitem[Ying et~al.(2019)Ying, Klein, Christiansen, Real, Murphy, and Hutter]{ying2019bench}
Chris Ying, Aaron Klein, Eric Christiansen, Esteban Real, Kevin Murphy, and Frank Hutter.
\newblock Nas-bench-101: Towards reproducible neural architecture search.
\newblock In \emph{International conference on machine learning}, pages 7105--7114. PMLR, 2019.

\bibitem[Zela et~al.(2019)Zela, Elsken, Saikia, Marrakchi, Brox, and Hutter]{zela2019understanding}
Arber Zela, Thomas Elsken, Tonmoy Saikia, Yassine Marrakchi, Thomas Brox, and Frank Hutter.
\newblock Understanding and robustifying differentiable architecture search.
\newblock \emph{arXiv preprint arXiv:1909.09656}, 2019.

\bibitem[Zhang and Jia(2021)]{zhang2021gradsign}
Zhihao Zhang and Zhihao Jia.
\newblock Gradsign: Model performance inference with theoretical insights.
\newblock \emph{arXiv preprint arXiv:2110.08616}, 2021.

\bibitem[Zhao et~al.(2021)Zhao, Dong, Shen, Zhang, Wei, and Chen]{zhao2021memory}
Yuekai Zhao, Li Dong, Yelong Shen, Zhihua Zhang, Furu Wei, and Weizhu Chen.
\newblock Memory-efficient differentiable transformer architecture search.
\newblock \emph{arXiv preprint arXiv:2105.14669}, 2021.

\bibitem[Zhou et~al.(2017)Zhou, Zhao, Puig, Fidler, Barriuso, and Torralba]{zhou2017scene}
Bolei Zhou, Hang Zhao, Xavier Puig, Sanja Fidler, Adela Barriuso, and Antonio Torralba.
\newblock Scene parsing through ade20k dataset.
\newblock In \emph{Proceedings of the IEEE conference on computer vision and pattern recognition}, pages 633--641, 2017.

\bibitem[Zoph et~al.(2018)Zoph, Vasudevan, Shlens, and Le]{nasnet}
Barret Zoph, Vijay Vasudevan, Jonathon Shlens, and Quoc~V Le.
\newblock Learning transferable architectures for scalable image recognition.
\newblock In \emph{Proceedings of the IEEE conference on computer vision and pattern recognition}, pages 8697--8710, 2018.

\end{thebibliography}
}

\end{document}